\documentclass[letterpaper, 10 pt, conference]{ieeeconf}

\IEEEoverridecommandlockouts
\usepackage{graphicx}
\usepackage{amsmath}
\usepackage{amssymb}
\usepackage{xspace}
\usepackage{adjustbox}
\usepackage{booktabs}
\usepackage{multirow}
\usepackage{tabularx}
\usepackage{multicol}           
\usepackage[table]{xcolor}
\usepackage{cite}
\usepackage{mathptmx}

\usepackage{algorithm}
\usepackage{algorithmic}
\usepackage{pifont}

\definecolor{AvgCol}{HTML}{EEF4FB}
\definecolor{RoboRow}{HTML}{EAF6F2}
\definecolor{RoboAvg}{HTML}{CFEAE1}
\definecolor{StrongGroup}{HTML}{A33D46}
\definecolor{WeakGroup}{HTML}{315F91}

\usepackage{hyperref}
\hypersetup{
    pdfauthor={Zixuan Chen, Lin Cao, Fenggang Zhang, Haotong Zhang, Yangtao Chen, Jieqi Shi, Jing Huo, Yang Gao},
    pdftitle={RoboLongR: Causally Consistent Rewards Teach Robots to Progressively Master Long-Horizon Manipulation},
    colorlinks=true,
    linkcolor=magenta,
    filecolor=magenta,
    urlcolor=magenta,
    citecolor=orange
}

\newcommand{\pname}{RoboLongR\xspace}

\title{\LARGE \bf
\pname: Causally Consistent Rewards Teach Robots to
Progressively Master Long-Horizon Manipulation
}

\author{
    Zixuan Chen$^{1}$,
    Lin Cao$^{1}$,
    Fenggang Zhang$^{1}$,
    Haotong Zhang$^{3}$,\\
    Yangtao Chen$^{2}$,
    Jieqi Shi$^{1}$,
    Jing Huo$^{1}$,
    and Yang Gao$^{1}$\\[0.5em]
    {
    $^{1}$Nanjing University \quad
    $^{2}$Fudan University \quad
    $^{3}$Southeast University
    }
}

\begin{document}

\maketitle
\thispagestyle{empty}
\pagestyle{empty}

\begin{abstract}
Long-horizon manipulation requires actions whose validity depends on earlier interactions, yet visual progress can conceal missing task prerequisites. We propose \textbf{RoboLongR}, an offline-to-online framework that learns history-dependent rewards to guide policy adaptation to new tasks. Its central objective is causal consistency: aligning rewarded progress with prerequisites supported by execution history. Stage-shuffle supervision encourages execution-order sensitivity, recurrent memory preserves evidence of earlier interactions, and two-round training refines trajectory segments using predicted progress. The resulting rewards guide stage-conditioned policy refinement from behavior cloning to online reinforcement learning. Sparse terminal feedback connects policy experience to reward learning by triggering correction for repeated high-scoring failures. Pretrained on demonstrations from 88 RLBench tasks and Open-X data, RoboLongR achieves 31.8\% average success on 18 long-horizon tasks excluded from reward pretraining, exceeding the strongest evaluated baseline by 9.2 percentage points. Across four real-robot tasks, it achieves 57.5\% success, a 32.5-percentage-point improvement over the comparison method. Reward-level controls probe sensitivity to synthetic order perturbations, while ablations examine the contributions of individual components to policy improvement.
Project website: \url{https://chenzixuan99.github.io/RoboLongR-public/}.
\end{abstract}
\begin{keywords}
Long-horizon manipulation, causally consistent rewards, reinforcement learning, learning from demonstration.
\end{keywords}

\section{Introduction}

Long-horizon manipulation requires robots to coordinate subtasks whose outcomes depend on earlier interactions~\cite{chen2026deco}. Demonstrations initialize behavior, while learned visual rewards supply dense feedback for further improvement through reinforcement learning (RL)~\cite{ma2023liv,sontakke2023roboclip,zhang2025rewind,luo2025hilserl,lei2025rl}. Transferable rewards extend this guidance to new tasks, allowing policies to improve beyond their demonstrations~\cite{zhang2025rewind,chen2026sarm,chen2026sarm2}. Reliable progress assessment is therefore central to expanding a robot's task repertoire through interaction.

\begin{figure}[t!]
    \centering
    \includegraphics[width=\linewidth]{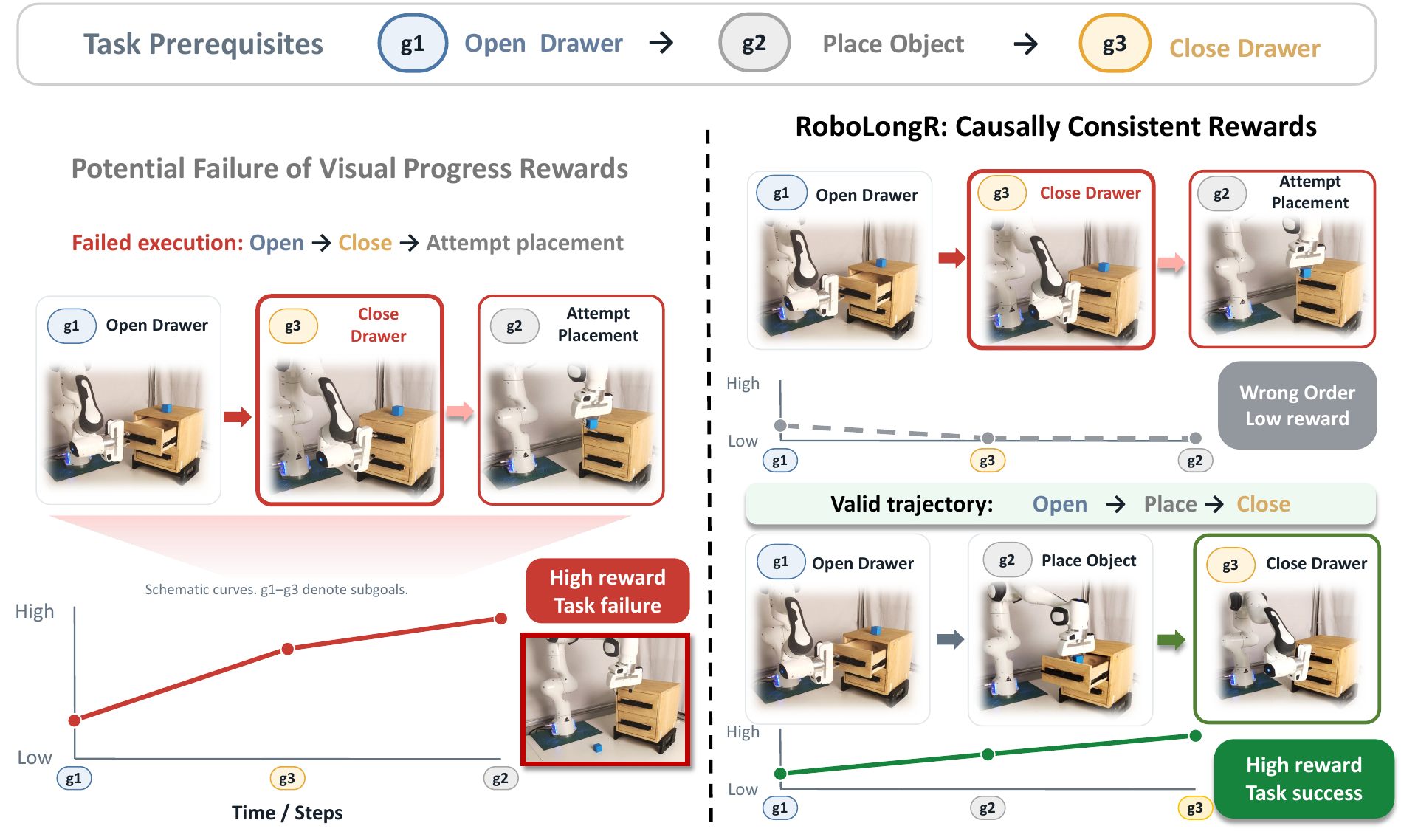}
    \caption{\textbf{Reward guidance beyond visual progress.} Closing an empty drawer can resemble task completion. RoboLongR combines stage-order supervision with execution history to encourage prerequisite-sensitive progress estimates. Labels $g_1$--$g_3$ denote semantic subgoals. Curves illustrate intended behavior, with responses on real executions evaluated separately.}
    \label{fig:teaser}
    \vspace{-5mm}
\end{figure}

However, recognizing a later task state does not establish that its prerequisites are satisfied. Drawer storage requires opening, object placement, and closing. Closing an empty drawer produces a goal-like image without completing the task (Fig.~\ref{fig:teaser}, left). Rewarding this apparent progress may reinforce skipped prerequisites. This ambiguity arises because identical observations can correspond to different outcomes depending on execution history. Rewards must therefore retain earlier interaction evidence, especially when online exploration leaves the demonstration distribution~\cite{pan2022misspecification,skalse2022defining}.

We call agreement between rewarded progress and task prerequisites supported by execution history \emph{causal consistency}. These prerequisites can form a partial order, allowing multiple valid sequences, and unsuccessful attempts can admit recovery. Learning such rewards poses two challenges: successful demonstrations provide little supervision for prerequisite violations, and adapting policies can encounter reward errors absent from training. This raises our central question: \emph{How can demonstrations and sparse interaction feedback yield rewards that guide adaptation to new tasks while remaining sensitive to task prerequisites?}

We propose \textbf{RoboLongR}, an offline-to-online framework that learns history-dependent rewards to guide progressive policy adaptation to new long-horizon tasks (Fig.~\ref{fig:teaser}, right). RoboLongR consists of three key components. First, to promote causal consistency, stage-shuffle reorders intact demonstration segments, providing execution-order supervision as a proxy for prerequisite sensitivity. Recurrent memory preserves evidence of earlier interactions, while a second training round refines segment boundaries using predicted progress. Second, RoboLongR uses the learned rewards to guide policy refinement beyond demonstrations. Target-task demonstrations initialize the policy through behavior cloning (BC), followed by reward-guided online RL. Coarse stages, defined independently of reward segmentation, condition both BC and scheduled RL. Every RL episode starts from a standard initial state and attempts the full task. Finally, to address reward errors encountered during policy adaptation, RoboLongR uses sparse terminal feedback to identify repeated high-scoring failures and trigger reward correction. This feedback loop allows the reward to guide policy learning while policy experience informs subsequent reward updates.

\begin{figure*}[t!]
    \centering
    \includegraphics[width=\textwidth]{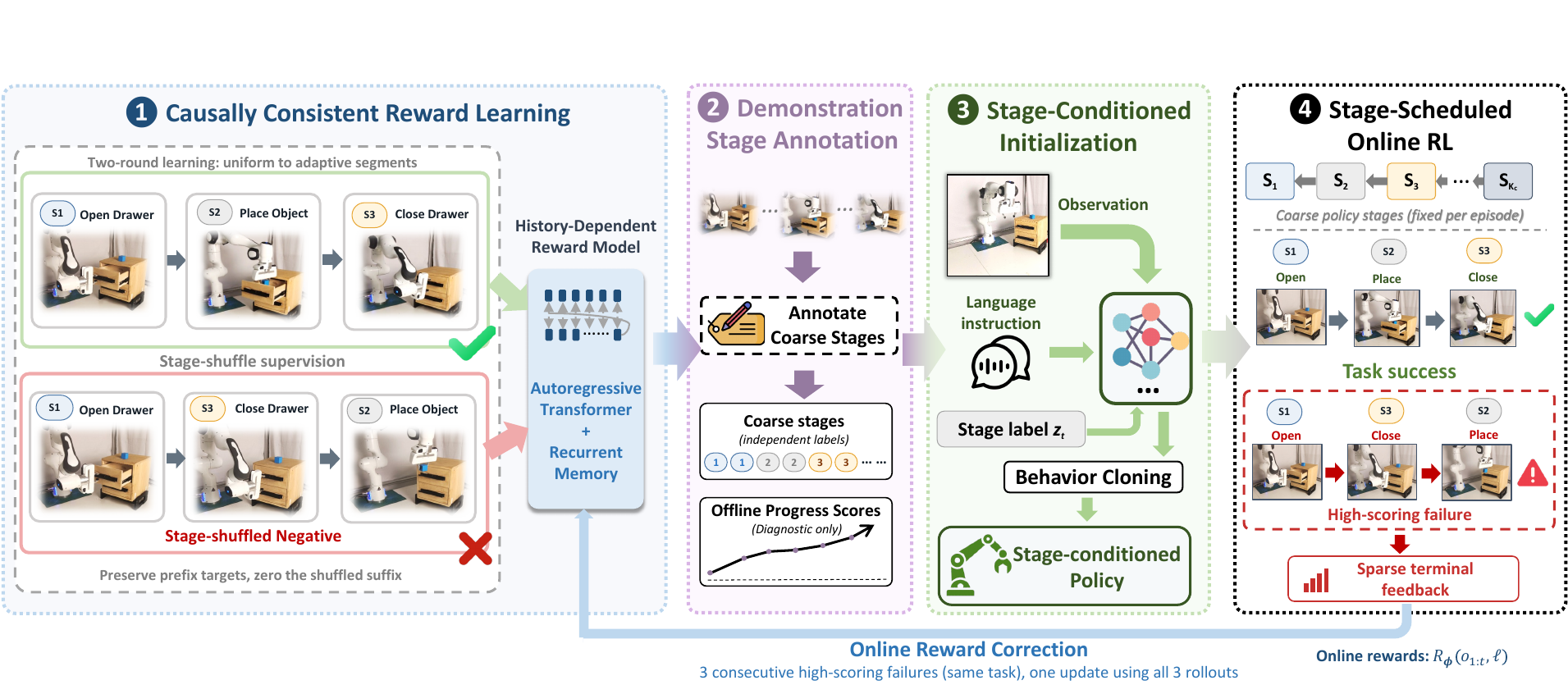}
    \caption{\textbf{RoboLongR connects reward learning and policy adaptation.} Reward-adaptive segments provide stage-order supervision. Separate coarse labels condition BC and scheduled online RL. Offline progress scores are diagnostic annotations. RL queries the reward on online histories, and sparse terminal feedback triggers reward correction after repeated high-scoring failures.}
    \label{fig:pipeline}
    \vspace{-5mm}
\end{figure*}

We pretrain the reward on demonstrations from 88 RLBench~\cite{james2020rlbench} tasks and Open-X Embodiment~\cite{oneill2024openx}, then evaluate policy adaptation on 18 disjoint long-horizon RLBench tasks. Target-task demonstrations initialize BC, and interaction supports policy refinement and reward correction. RoboLongR achieves 31.8\% average success, compared with 7.8\% for BC alone, and exceeds the strongest evaluated baseline by 9.2 percentage points. On four real-robot tasks, it achieves 57.5\% success, a 32.5-percentage-point improvement over the comparison method.

In summary, our contributions are threefold. (1) We combine stage-order supervision and recurrent memory to learn progress rewards sensitive to execution history and task prerequisites. (2) We develop a reward-guided BC-to-RL framework that separates reward segmentation from policy conditioning and uses sparse terminal feedback to correct recurring reward errors. (3) We evaluate policy adaptation on tasks excluded from reward pretraining and on real robots, supported by reward-level controls and component ablations.

\section{Related Work}
\noindent\textbf{Rewards for Robot Learning.}
Learning rewards from demonstrations has been studied through inverse reinforcement learning and adversarial imitation learning~\cite{abbeel2004irl,ho2016gail}. Recent methods use visual and language representations to estimate progress, value, or similarity for robot control, including LIV~\cite{ma2023liv}, VLC~\cite{alakuijala2024vlc}, GVL~\cite{ma2024gvl}, and RoboCLIP~\cite{sontakke2023roboclip}. ReWiND~\cite{zhang2025rewind} learns transferable progress rewards from rewound trajectories for offline-to-online policy adaptation. General-purpose reward models further explore trajectory comparisons~\cite{liang2026robometer}, outcome assessment~\cite{lee2026roboreward}, and dense process supervision~\cite{zhai2025vision}. These approaches broaden the signals available for policy learning, but estimated progress alone does not establish that earlier prerequisites were satisfied. RoboLongR focuses on this distinction, combining stage-order supervision with recurrent execution memory to encourage prerequisite-sensitive reward learning.

\noindent\textbf{Stage-Aware Rewards and Policy Refinement.}
Stage modeling structures reward estimation in long-horizon tasks. SARM~\cite{chen2026sarm} jointly predicts stages and fine-grained progress, while SARM2~\cite{chen2026sarm2} extends stage-aware rewards to multiple tasks and improvement from autonomous rollouts. RoboHorizon~\cite{chen2025robohorizon} combines LLM-generated subtask rewards with a multi-view world model. As policies improve through interaction~\cite{luo2025hilserl,lei2025rl,yang2026rise,liu2026lifelongrft,zhang2025robustvla}, changing behavior can expose unreliable reward estimates. RARM~\cite{yang2026rarm} addresses this issue through confidence gating. RoboLongR combines distinct stage representations with feedback-driven correction: adaptive segments construct order supervision, coarse stages condition the policy, and sparse terminal outcomes identify repeated high-scoring failures. Correcting the reward on these failures connects subsequent guidance to evidence gathered during policy execution.

\section{Method}
\label{sec:method}
RoboLongR learns a reward from multitask demonstrations to guide downstream policy adaptation. The framework comprises four stages (Fig.~\ref{fig:pipeline}): causally consistent reward learning, demonstration stage annotation, stage-conditioned initialization, and stage-scheduled online RL. Terminal feedback connects policy experience to reward learning by identifying recurring high-scoring failures.

\subsection{Problem Formulation}
Given a task instruction $\ell$, let $\tau=\{(o_t,s_t,a_t)\}_{t=1}^{T}$ denote an execution, where $o_t$, $s_t$, and $a_t$ are RGB observations, robot states, and actions. The environment provides binary terminal success $b(\tau)$. Our goal is to learn a dense progress reward $R_\phi(o_{1:t},\ell)\in[0,1]$ that supports policy improvement beyond the demonstrations.

Causal consistency aligns rewarded progress with prerequisites supported by the observed history. These prerequisites may form a partial order. Without annotated prerequisite graphs, we pursue this objective through stage-order supervision and terminal-feedback correction.

\subsection{Causally Consistent Reward Learning}
\label{sec:reward_learning}
\noindent\textbf{Stage-shuffle supervision.}
Successful demonstrations rarely show familiar actions executed without their prerequisites. To create order mismatches, we partition each demonstration into contiguous segments $S_1,\ldots,S_{K_a(\tau)}$ and sample a shuffled counterpart with probability $p=0.5$. For three consecutive segments, stage-shuffle replaces
$S_i\!\to\!S_{i+1}\!\to\!S_{i+2}$
with
$S_i\!\to\!S_{i+2}\!\to\!S_{i+1}$.
Each segment retains its visual content and internal ordering in a different execution context. For example, opening--closing--placement violates the demonstrated drawer sequence.

Let $t_*$ be the first displaced step. We preserve the positive progress target before $t_*$ and set the remaining targets to zero:
\begin{equation}
\tilde y_t=
\begin{cases}
y_t, & t<t_*,\\
0, & t\geq t_*.
\end{cases}
\end{equation}
Padding-masked MSE supervises original and sampled shuffled prefixes using $y_t$ and $\tilde y_t$, respectively. Original positives remain in training even when no shuffle is sampled.

Order-preserving splices provide training positives and a control on disjoint held-out demonstrations (Sec.~\ref{sec:q2}). Their boundary statistics are not exactly matched to shuffles, and zero-suffix supervision can penalize valid alternative orders or recoveries.

\noindent\textbf{Recurrent execution memory.}
Order-sensitive prediction requires evidence of interactions that may no longer be visible. Frozen DINOv2 ViT-B/14~\cite{oquab2024dinov2} encodes two RGB views, and frozen all-MiniLM-L6-v2 encodes the instruction. Projected features pass through an autoregressive Transformer, a recurrent module, and an MLP progress head. Attention spans 100 frames. Longer sequences use causal chunks, carrying recurrent state across chunks and detaching it during training. Memory resets between episodes. Each prediction uses only its observed prefix, while recurrent memory carries information beyond the attention window. Appendix~\ref{app:reward_inputs} details view aggregation and recurrent-state initialization.

\noindent\textbf{Reward-adaptive segmentation.}
To adapt shuffles to progress changes, we refine segments in two rounds. Round 1 uses uniform partitions and targets $y_t=(t/T)^2$. Round 2 continues from the first-round checkpoint, adapting segments to progress changes and using per-demonstration min-max normalized predictions as targets.
For adaptive partitioning, we smooth each predicted curve with an 11-frame, degree-three Savitzky--Golay filter. Boundaries are local maxima of the absolute second derivative that exceed one standard deviation of curvature magnitude and are separated by at least five frames. If no boundary qualifies, we use five uniform segments. Stage-shuffle requires at least three segments. These partitions follow predicted progress and need not coincide with semantic subtasks. Appendix~\ref{app:adaptive_partition} reports partition frequency and granularity across task categories.

\noindent\textbf{Data augmentation.}
Each base batch contains six RLBench and two Open-X trajectories. We retain all originals and generate augmented variants in separate branches without composing augmentations.

\emph{Rewind.} With probability $0.5$ per Open-X trajectory, we uniformly sample an integer pivot $i\in[\lceil0.4T\rceil,\lfloor0.8T\rfloor]$, followed by a rewind length $k\in[\lceil0.1T\rceil,\min(\lfloor0.3T\rfloor,i-1,T-i)]$. Cases with empty sampling ranges are skipped. Following ReWiND~\cite{zhang2025rewind}, we construct $(o_{1:i},o_{i-1},\ldots,o_{i-k})$. Each frame inherits its source-frame target $y_{h(t)}$, where $h(t)=t$ for $t\leq i$ and $h(t)=2i-t$ otherwise. Both training rounds retain these targets without renormalizing by augmented length or assigning zero targets to the reversed suffix.

\emph{Splice.} With probability $0.25$ per RLBench trajectory, we replace one semantic stage with a matching stage from another successful demonstration in the same split. We uniformly sample an eligible stage and donor with matching task variant, subtask, and object role, compatible pre- and postconditions, and a donor-to-recipient length ratio in $[0.8,1.25]$. Matching requires semantic annotations and compatibility checks beyond adaptive segment indices. Cases without a valid match are skipped. Uniform temporal resampling with nearest-neighbor selection matches the recipient length, while recipient targets are retained in both rounds. Both views share sampling indices. Recurrent memory resets at the start of each augmented trajectory and carries across splice boundaries.

For each original or augmented trajectory, $L_j^a$ denotes the mean squared error over valid, unpadded frames. The reward-training objective is
\begin{equation}
\label{eq:reward_training}
\mathcal L_R=\frac18\!\left[
\sum_{j=1}^{8}L_j^+
+\sum_{j\in\mathcal S}L_j^{\rm sh}
+\sum_{j\in\mathcal R}L_j^{\rm rw}
+0.5\!\sum_{j\in\mathcal P}L_j^{\rm sp}
\right],
\end{equation}
where $\mathcal S$, $\mathcal R$, and $\mathcal P$ index the sampled shuffle, rewind, and splice variants, respectively. Normalizing by the eight originals preserves the effects of augmentation sampling rates. Algorithm~\ref{alg:reward_training} summarizes the training procedure.

\begin{algorithm}[t]
\caption{Proposed two-round reward pretraining}
\label{alg:reward_training}
\begin{algorithmic}[1]
\REQUIRE RLBench/Open-X demonstrations and semantic splice matches
\STATE initialize $R_\phi$
\FOR{$q=1,2$}
  \STATE set round-specific targets and RLBench partitions
  \FOR{each base batch of six RLBench and two Open-X trajectories}
    \STATE retain original positives and sample separate shuffle ($0.5$), rewind ($0.5$), and splice ($0.25$) branches
    \STATE assign prefix/zero-suffix, source-frame, or recipient targets, respectively
    \STATE update $\phi$ using Eq.~\ref{eq:reward_training}
  \ENDFOR
\ENDFOR
\RETURN $R_\phi$
\end{algorithmic}
\end{algorithm}

\subsection{Demonstration Stage Annotation}
\label{sec:stage_annotation}
Policy conditioning requires shared labels across demonstrations. We assign coarse labels $z_t\in\{1,\ldots,K_c\}$ using task-category templates, such as reach, grasp, lift, transport, and place. Gripper transitions, object motion, contact, and release determine boundaries. The template size $K_c$ is independent of the trajectory-specific $K_a(\tau)$. Coarse labels condition BC and RL, while adaptive segments structure reward supervision. Offline scores are diagnostic annotations. BC uses expert actions and coarse labels, and PPO queries online histories.

\subsection{Stage-Conditioned Initialization}
\label{sec:policy_initialization}
BC initializes the policy using expert actions and per-step labels for 50 epochs. The actor and critic condition on $(f_t,s_t,\mathbf e_\ell,z_t)$, where $f_t$ projects mean-pooled frozen DINOv2 features from two views to 512 dimensions, separately from the reward network. Robot state includes joint positions and velocities, end-effector pose, and gripper opening. Language features are frozen, and stage embeddings are learned. Actions comprise end-effector displacement, quaternion rotation, and a gripper command. The policy uses current observations, while reward prediction uses execution history.

\subsection{Stage-Scheduled Online RL}
\label{sec:online_rl}
PPO~\cite{schulman2017ppo} refines the BC policy using the learned progress score directly as its per-step reward: $r_t^{\rm RL}=R_\phi(o_{1:t},\ell).$
PPO optimizes $\mathbb E_\pi[\sum_{t=1}^{T}\gamma^{t-1}r_t^{\rm RL}]$ with $\gamma=0.99$. Episodes end at success, terminal failure, or the 500-step horizon, without a terminal bonus. GAE bootstraps only at nonterminal rollout cuts. Progress return is a success surrogate that can favor prolonged high-scoring unfinished executions over early completion. Terminal feedback targets recurring failures but does not eliminate this incentive by construction.

Stage inputs select BC conditioning modes and remain fixed within full-task episodes starting from ordinary initial states. The schedule starts at $K_c$. At $k>1$, at least 17 successes in a rolling window of $W=20$ completed episodes trigger $k\to k-1$ for the next episode and reset the window. Stage 1 remains active once reached. Training ends at the interaction budget without forced transitions, so runs need not visit all stages. Evaluation always uses stage 1.

\noindent\textbf{Online reward correction.}
To detect recurring mismatches between reward and task success, we flag a rollout when $b(\tau)=0$ and $\max_t R_\phi(o_{1:t},\ell)>0.65$. Any non-flagged episode clears that task's consecutive-failure buffer. Three consecutive flagged episodes for the same task form a batch $\mathcal B^-=\{\tau_j^-\}_{j=1}^{3}$. Correction uses all three trajectories through
\begin{equation}
\mathcal L_{\rm corr}=\frac{1}{3}\sum_{j=1}^{3}
\frac{1}{T_j}\sum_{t=1}^{T_j}R_\phi(o_{1:t}^{(j)},\ell)^2,
\end{equation}
where $T_j$ and $o_{1:t}^{(j)}$ denote the length and observation prefix of $\tau_j^-$. One Adam step updates the Transformer, recurrent module, and progress head with learning rate $10^{-4}$ and gradient norm clipped to 1.0. The encoders remain frozen. Each completed triple resets the trigger buffer. Correction uses no offline replay.

Reward parameters remain frozen throughout each PPO collection and optimization cycle. Correction uses queued failure triples after the PPO update, leaving collected rewards and advantages unchanged. Subsequent collection uses the updated model. For episodes spanning an update, the model replays the observed prefix to rebuild reward memory and refresh the trigger maximum.

Policy experience thus revises reward guidance without new per-step labels. However, terminal failure cannot localize an error, so whole-trajectory correction can suppress valid partial progress.

\noindent\textbf{Implementation details.}
Visual and language features have dimensions 768 and 384, both projected to 512. Images use $224\times224$ crops with DINOv2 normalization. The Transformer has four layers, eight heads, feed-forward width 2048, and dropout 0.1, followed by 512-dimensional recurrent memory. Reward learning uses Adam with learning rate $10^{-4}$, batch size 8, and 60 epochs in total, with 30 per round. PPO uses learning rate $3\times10^{-4}$, clip ratio 0.2, GAE 0.95, 2,048-step rollouts, and ten optimization epochs. Budgets are $2\times10^5$ steps per simulation task and 120 episodes per real-robot task at 10 Hz. Simulation uses seeds 0, 1, and 2 on eight RTX 4090 GPUs. Appendix~\ref{app:randomization} describes randomization and hyperparameter selection.

\begin{figure}[!t]
    \centering
    \includegraphics[width=\linewidth]{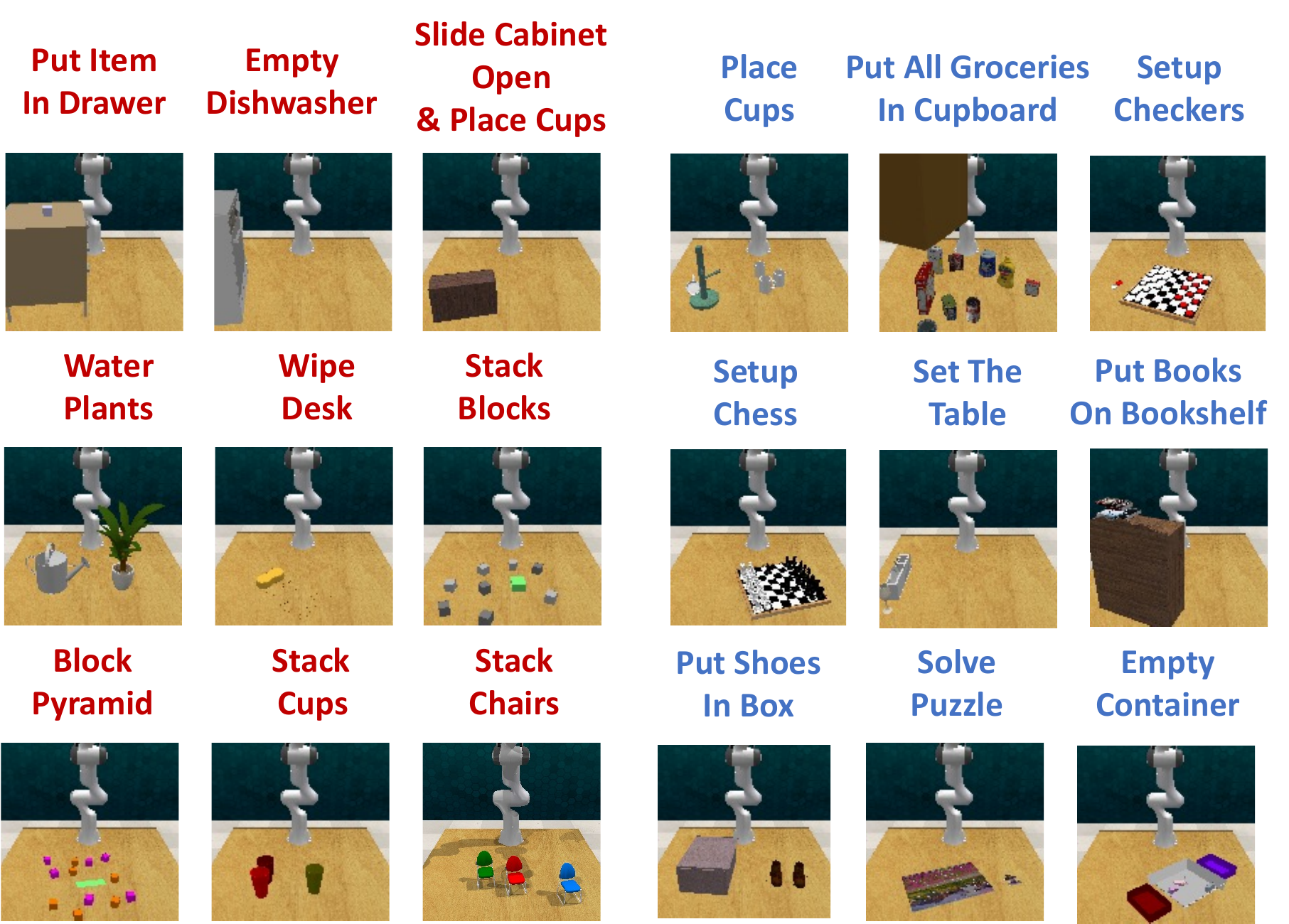}
\caption{\small \textbf{RLBench evaluation tasks.} The 18 long-horizon tasks are excluded from offline reward pretraining.}
    \label{fig:sim_tasks}
    \vspace{-4mm}
\end{figure}

\begin{table*}[!t]
\centering
\caption{\small \textbf{Success rates (\%) on 18 long-horizon RLBench tasks grouped by dependency strength.} Results report mean $\pm$ sample standard deviation over three seeds, with 50 trials per seed. Avg. denotes the within-group mean.}
\label{tab:main_results}
\resizebox{\textwidth}{!}{%
\scriptsize
\setlength{\tabcolsep}{3pt}
\begin{tabular}{l>{\columncolor{AvgCol}}c*{9}{c}}
\toprule
\multicolumn{11}{c}{\textcolor{StrongGroup}{\textit{\textbf{Strong-dependency tasks with prominent physical prerequisites}}}} \\
\midrule
Models
& \begin{tabular}[b]{@{}c@{}}Avg.\\(\%) $\uparrow$\end{tabular}
& \begin{tabular}[b]{@{}c@{}}Put Item\\In Drawer\end{tabular}
& \begin{tabular}[b]{@{}c@{}}Empty\\Dishwasher\end{tabular}
& \begin{tabular}[b]{@{}c@{}}Slide Cabinet Open\\\& Place Cups\end{tabular}
& \begin{tabular}[b]{@{}c@{}}Water\\Plants\end{tabular}
& \begin{tabular}[b]{@{}c@{}}Wipe\\Desk\end{tabular}
& \begin{tabular}[b]{@{}c@{}}Block\\Pyramid\end{tabular}
& \begin{tabular}[b]{@{}c@{}}Stack\\Blocks\end{tabular}
& \begin{tabular}[b]{@{}c@{}}Stack\\Cups\end{tabular}
& \begin{tabular}[b]{@{}c@{}}Stack\\Chairs\end{tabular} \\
\midrule
BC-only
& 8.2
& 14.0\tiny{$\pm$3.5}
& 9.3\tiny{$\pm$2.3}
& 6.7\tiny{$\pm$3.1}
& 16.7\tiny{$\pm$3.1}
& 20.0\tiny{$\pm$2.0}
& 0.0
& 2.7\tiny{$\pm$1.2}
& 4.7\tiny{$\pm$4.2}
& 0.0 \\
Sparse-PPO
& 7.1
& 12.7\tiny{$\pm$2.3}
& 8.0\tiny{$\pm$3.5}
& 5.3\tiny{$\pm$2.3}
& 14.7\tiny{$\pm$4.2}
& 18.0\tiny{$\pm$3.5}
& 0.0
& 2.0\tiny{$\pm$2.0}
& 3.3\tiny{$\pm$3.1}
& 0.0 \\
LIV
& 3.5
& 6.7\tiny{$\pm$4.2}
& 4.0\tiny{$\pm$3.5}
& 1.3\tiny{$\pm$2.3}
& 8.0\tiny{$\pm$3.5}
& 11.3\tiny{$\pm$1.2}
& 0.0
& 0.0
& 0.0
& 0.0 \\
Progress+Mono
& 20.9
& 30.7\tiny{$\pm$3.1}
& 24.0\tiny{$\pm$2.0}
& 20.0\tiny{$\pm$2.0}
& 33.3\tiny{$\pm$4.2}
& 38.0\tiny{$\pm$2.0}
& 8.7\tiny{$\pm$3.1}
& 11.3\tiny{$\pm$4.2}
& 15.3\tiny{$\pm$3.1}
& 6.7\tiny{$\pm$4.2} \\
ReWiND
& 23.6
& 34.0\tiny{$\pm$3.5}
& 27.3\tiny{$\pm$2.3}
& 22.7\tiny{$\pm$4.2}
& 36.7\tiny{$\pm$4.2}
& 41.3\tiny{$\pm$1.2}
& 10.0\tiny{$\pm$2.0}
& 14.0\tiny{$\pm$2.0}
& 17.3\tiny{$\pm$4.2}
& 8.7\tiny{$\pm$2.3} \\
\rowcolor{RoboRow}
\textbf{RoboLongR}
& \cellcolor{RoboAvg}\textbf{37.8}
& \textbf{49.3}\tiny{$\pm$2.3}
& \textbf{42.0}\tiny{$\pm$2.0}
& \textbf{36.7}\tiny{$\pm$2.3}
& \textbf{51.3}\tiny{$\pm$2.3}
& \textbf{55.3}\tiny{$\pm$3.1}
& \textbf{22.7}\tiny{$\pm$3.1}
& \textbf{27.3}\tiny{$\pm$4.2}
& \textbf{32.0}\tiny{$\pm$3.5}
& \textbf{23.3}\tiny{$\pm$4.2} \\
\midrule
\midrule
\multicolumn{11}{c}{\textcolor{WeakGroup}{\textit{\textbf{Weak-dependency tasks allowing flexible subtask order}}}} \\
\midrule
Models
& \begin{tabular}[b]{@{}c@{}}Avg.\\(\%) $\uparrow$\end{tabular}
& \begin{tabular}[b]{@{}c@{}}Place\\Cups\end{tabular}
& \begin{tabular}[b]{@{}c@{}}Put All Groceries\\In Cupboard\end{tabular}
& \begin{tabular}[b]{@{}c@{}}Setup\\Checkers\end{tabular}
& \begin{tabular}[b]{@{}c@{}}Setup\\Chess\end{tabular}
& \begin{tabular}[b]{@{}c@{}}Set The\\Table\end{tabular}
& \begin{tabular}[b]{@{}c@{}}Put Books\\On Bookshelf\end{tabular}
& \begin{tabular}[b]{@{}c@{}}Put Shoes\\In Box\end{tabular}
& \begin{tabular}[b]{@{}c@{}}Solve\\Puzzle\end{tabular}
& \begin{tabular}[b]{@{}c@{}}Empty\\Container\end{tabular} \\
\midrule
BC-only
& 7.4
& 16.0\tiny{$\pm$4.0}
& 2.0\tiny{$\pm$3.5}
& 3.3\tiny{$\pm$1.2}
& 0.0
& 0.0
& 12.7\tiny{$\pm$4.2}
& 10.0\tiny{$\pm$4.0}
& 0.0
& 22.7\tiny{$\pm$4.2} \\
Sparse-PPO
& 6.5
& 14.0\tiny{$\pm$3.5}
& 1.3\tiny{$\pm$1.2}
& 2.7\tiny{$\pm$2.3}
& 0.0
& 0.0
& 11.3\tiny{$\pm$3.1}
& 8.7\tiny{$\pm$2.3}
& 0.0
& 20.7\tiny{$\pm$2.3} \\
LIV
& 2.4
& 6.0\tiny{$\pm$4.0}
& 0.0
& 0.0
& 0.0
& 0.0
& 4.7\tiny{$\pm$1.2}
& 4.0\tiny{$\pm$3.5}
& 0.0
& 6.7\tiny{$\pm$2.3} \\
Progress+Mono
& 19.3
& 30.7\tiny{$\pm$2.3}
& 10.7\tiny{$\pm$3.1}
& 13.3\tiny{$\pm$4.2}
& 6.7\tiny{$\pm$1.2}
& 5.3\tiny{$\pm$3.1}
& 26.0\tiny{$\pm$2.0}
& 22.7\tiny{$\pm$3.1}
& 7.3\tiny{$\pm$2.3}
& 50.7\tiny{$\pm$4.2} \\
ReWiND
& 21.6
& 34.0\tiny{$\pm$2.0}
& 12.7\tiny{$\pm$4.2}
& 15.3\tiny{$\pm$2.3}
& 8.0\tiny{$\pm$2.0}
& 6.7\tiny{$\pm$3.1}
& 29.3\tiny{$\pm$1.2}
& 26.0\tiny{$\pm$2.0}
& 8.7\tiny{$\pm$2.3}
& 54.0\tiny{$\pm$3.5} \\
\rowcolor{RoboRow}
\textbf{RoboLongR}
& \cellcolor{RoboAvg}\textbf{25.9}
& \textbf{38.0}\tiny{$\pm$2.0}
& \textbf{16.7}\tiny{$\pm$4.2}
& \textbf{19.3}\tiny{$\pm$2.3}
& \textbf{11.3}\tiny{$\pm$4.2}
& \textbf{10.0}\tiny{$\pm$4.0}
& \textbf{33.3}\tiny{$\pm$2.3}
& \textbf{34.0}\tiny{$\pm$2.0}
& \textbf{12.7}\tiny{$\pm$3.1}
& \textbf{57.3}\tiny{$\pm$3.1} \\
\bottomrule
\end{tabular}}

\end{table*}

\section{Experiments}
We evaluate RoboLongR through four questions. (1) Can the learned rewards improve policies on tasks excluded from reward pretraining (Sec.~\ref{sec:main_results})? (2) Are the rewards sensitive to changes in execution order (Sec.~\ref{sec:q2})? (3) Which components contribute to policy improvement (Sec.~\ref{sec:ablation})? (4) Do the gains extend to real robots, and how do rewards respond to prerequisite-violating executions (Sec.~\ref{sec:real_robot})?

\subsection{Experimental Setup}
\label{sec:experimental_setup}
\label{sec:setup}

\noindent\textbf{Benchmark and data.}
We separate reward pretraining from policy adaptation using 88 RLBench~\cite{james2020rlbench} pretraining tasks and 18 disjoint long-horizon evaluation tasks (Fig.~\ref{fig:sim_tasks}). Appendix~\ref{app:sim_tasks} lists the pretraining tasks and evaluation-task prerequisites. 
Pretraining uses 20 successful demonstrations per task, totaling 1,760 trajectories, and 4,000 trajectories from the bridge and fractal20220817 subsets of Open-X Embodiment~\cite{oneill2024openx}. Open-X contributes 25\% of each base batch, supplying progress supervision and rewound negatives. RLBench provides stage-shuffle supervision for the shared reward model. ReWiND is retrained on the same source-data mixture with its objective.
Each evaluation task provides 20 demonstrations for stage annotation and BC initialization. Online interaction then supports policy refinement and reward correction. We group the tasks into nine \textcolor{StrongGroup}{\textbf{strong-dependency tasks}} with prominent physical prerequisites and nine \textcolor{WeakGroup}{\textbf{weak-dependency tasks}} allowing flexible subtask order. Drawer storage requires opening before object placement, whereas table setting permits independent placements. Both groups can contain partial orders. This grouping examines how gains vary with task structure, while ablations assess individual component contributions.

\noindent\textbf{Baselines.}
All RL methods share demonstrations, stage-conditioned policy architecture, BC initialization, interaction budgets, and stage schedule. \textit{BC-only} evaluates the initialized policy, and \textit{Sparse-PPO} uses binary terminal rewards. \textit{LIV}~\cite{ma2023liv} supplies pretrained rewards without task-specific reward training, min-max normalized per rollout before PPO. \textit{ReWiND}~\cite{zhang2025rewind} learns progress from rewound negatives. \textit{Progress+Mono} shares RoboLongR's reward backbone, replacing stage-shuffle with stage prediction and a penalty on decreasing expected stage indices. Its objective combines positive progress regression, stage cross-entropy, and monotonicity, with unit weights on both auxiliary terms (Appendix~\ref{app:baselines}). Online reward correction applies to RoboLongR and Progress+Mono, while other baselines use fixed reward functions. The main comparison evaluates complete learning pipelines, with component contributions examined separately by ablation.

\noindent\textbf{Evaluation protocol.}
We evaluate complete-task success independently of learned return. Each simulation task uses three seeds and 50 evaluation episodes per seed with randomized layouts, totaling 150 trials. Results report mean success and sample standard deviation across seeds, with benchmark averages weighting tasks equally.

\subsection{Policy Improvement on RLBench}
\label{sec:main_results}

\noindent\textbf{Reward-guided adaptation.}
RoboLongR improves mean success from the BC initialization of 7.8\% to 31.8\% on the 18 held-out tasks (Table~\ref{tab:main_results}). Under the same demonstration and interaction budgets, it exceeds ReWiND, the strongest evaluated baseline at 22.6\%, by 9.2 percentage points. Sparse-PPO remains near BC at 6.8\%, while LIV and Progress+Mono achieve 2.9\% and 20.1\%. These results support reward-guided policy adaptation to tasks excluded from reward pretraining. The gap to Progress+Mono further indicates that stage prediction and monotonicity do not reproduce the full framework's gains.

\noindent\textbf{Dependence on task structure.}
RoboLongR achieves higher mean success than ReWiND on all 18 tasks. Group averages are 37.8\% versus 23.6\% for strong-dependency tasks and 25.9\% versus 21.6\% for weak-dependency tasks. Computed before rounding, the gains are 14.2 and 4.2 percentage points. The larger improvement on strong-dependency tasks is consistent with the motivation for history-sensitive rewards, although the groups also differ in manipulation demands. The following reward diagnostics and ablations examine execution-order sensitivity and component contributions more directly.

\subsection{Execution-Order Sensitivity}
\label{sec:q2}

\noindent\textbf{Reward-level protocol.}
To assess execution-history sensitivity independently of policy performance, we evaluate 180 correct demonstrations, ten additional per evaluation task, excluded from reward training, BC, and model selection. Each condition contains 180 correct--perturbed pairs scored by one final frozen pretraining checkpoint per method. Order perturbations swap intact segments, rewound trajectories introduce temporal regression, and order-preserving splices replace a segment with the corresponding stage from another successful demonstration. Appendix~\ref{app:permutation_distance} details the swap-distance groups. Valid orders observed in held-out weak-dependency demonstrations remain in the correct set. All variants remain held out. We compute raw mean rewards $\bar r(\tau)=T^{-1}\sum_t R_\phi(o_{1:t},\ell)$ and paired margins $\Delta_x=\bar r(\tau)-\bar r(\tau^x)$, weighting pairs equally. Ranking compares correct/order-perturbed pairs, counting ties as incorrect. Progress MSE averages per-trajectory errors against $(t/T)^2$ on correct demonstrations.

\begin{table}[!t]
\centering
\caption{\small \textbf{Reward-level diagnostics on held-out trajectories.} $\Delta$ denotes the average paired difference in mean trajectory reward.}
\label{tab:causal_sensitivity}
\small
\begin{tabular}{lcc}
\toprule
 & ReWiND & RoboLongR \\
\midrule
$\Delta$(order-perturbed) $\uparrow$ & 0.04 & \textbf{0.40} \\
$\Delta$(rewound) $\uparrow$ & 0.36 & 0.39 \\
$\Delta$(spliced), near $0$ & 0.03 & 0.05 \\
Ranking accuracy (\%) $\uparrow$ & 55.0 & \textbf{90.0} \\
Progress MSE $\downarrow$ & 0.021 & 0.023 \\
\bottomrule
\end{tabular}
\vspace{-2mm}
\end{table}

\noindent\textbf{Sensitivity and controls.}
RoboLongR separates correct and order-perturbed trajectories more clearly than ReWiND (Table~\ref{tab:causal_sensitivity}), with ranking accuracies of 90.0\% versus 55.0\% and paired margins of 0.40 versus 0.04. Both methods distinguish rewound trajectories, with margins of 0.39 and 0.36. RoboLongR's margin of 0.05 for order-preserving splices is substantially smaller than its order-perturbation margin. Progress MSEs remain close at 0.023 and 0.021. These results support sensitivity to the tested order perturbations, with comparable progress prediction on correct demonstrations.

\subsection{Ablation Studies}
\label{sec:ablation}

\begin{figure}[!t]
\centering
\includegraphics[width=\columnwidth]{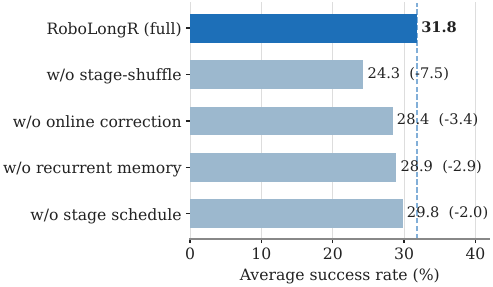}
\caption{\small \textbf{Component ablations on RLBench.} Mean task success after removing individual components under shared demonstrations and interaction budgets.}
\label{fig:ablation}
\vspace{-4mm}
\end{figure}

\begin{figure}[!t]
    \centering
    \includegraphics[width=0.9\columnwidth]{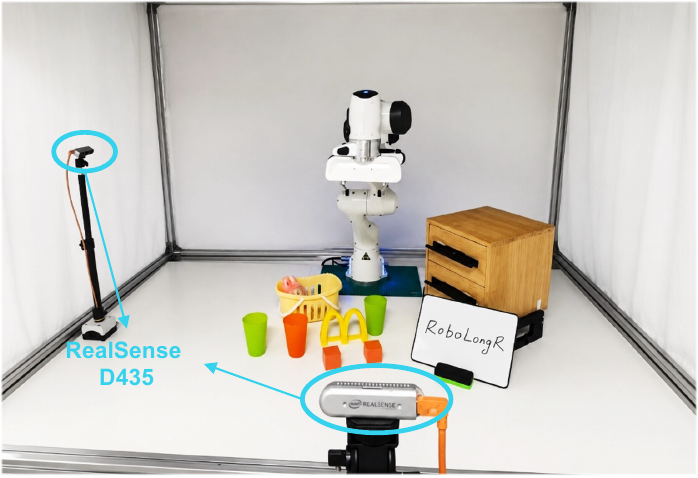}
\caption{\small \textbf{Real-robot platform.} A Franka Panda with front and right RealSense D435 cameras.}
    \label{fig:real_setup}
    \vspace{-4mm}
\end{figure}

\begin{figure}[!t]
    \centering
    \includegraphics[width=\columnwidth]{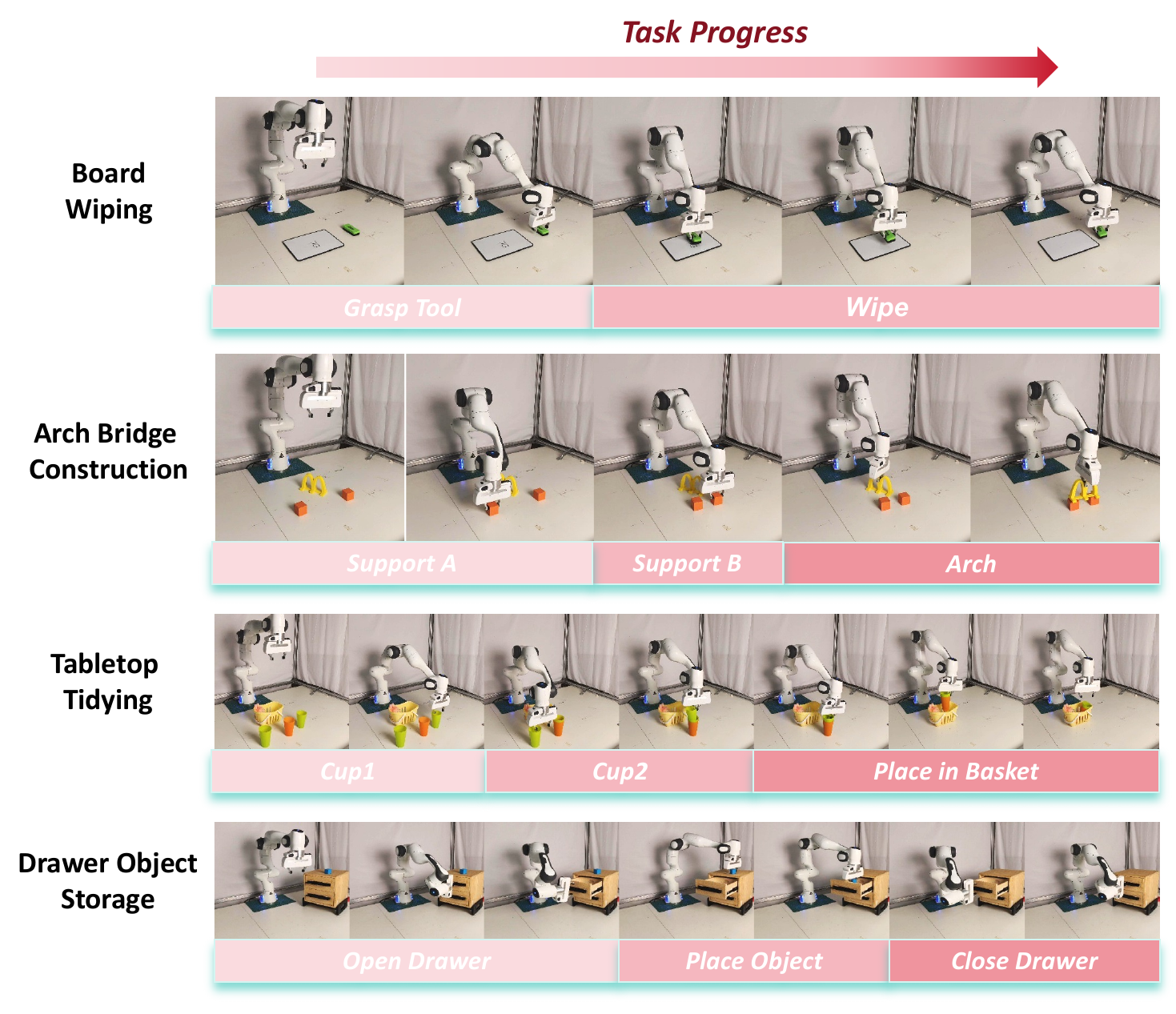}
    \caption{\small \textbf{Four real-robot tasks and their representative subtask sequences.}}
    \label{fig:real_tasks}
    \vspace{-2mm}
\end{figure}

\begin{figure*}[!t]
    \centering
    \includegraphics[width=0.98\textwidth]{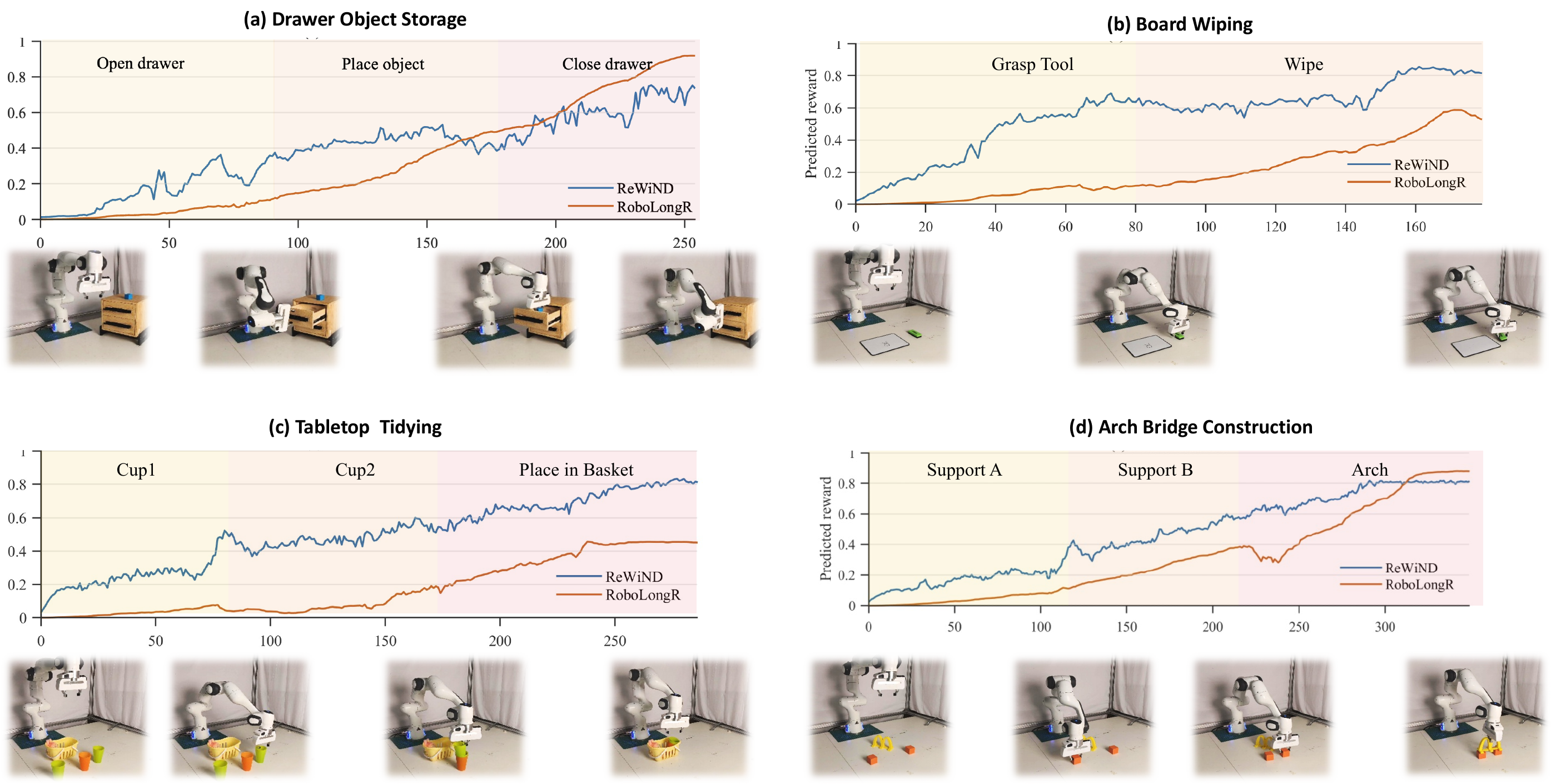}
    \caption{\textbf{Reward progress on valid real-robot executions.} Shaded regions indicate semantic task stages, and images show corresponding observations. Raw reward scales are model-specific.}
    \label{fig:real_valid_reward}
    \vspace{-2mm}
\end{figure*}

\begin{figure}[!t]
    \centering
    \includegraphics[width=\columnwidth]{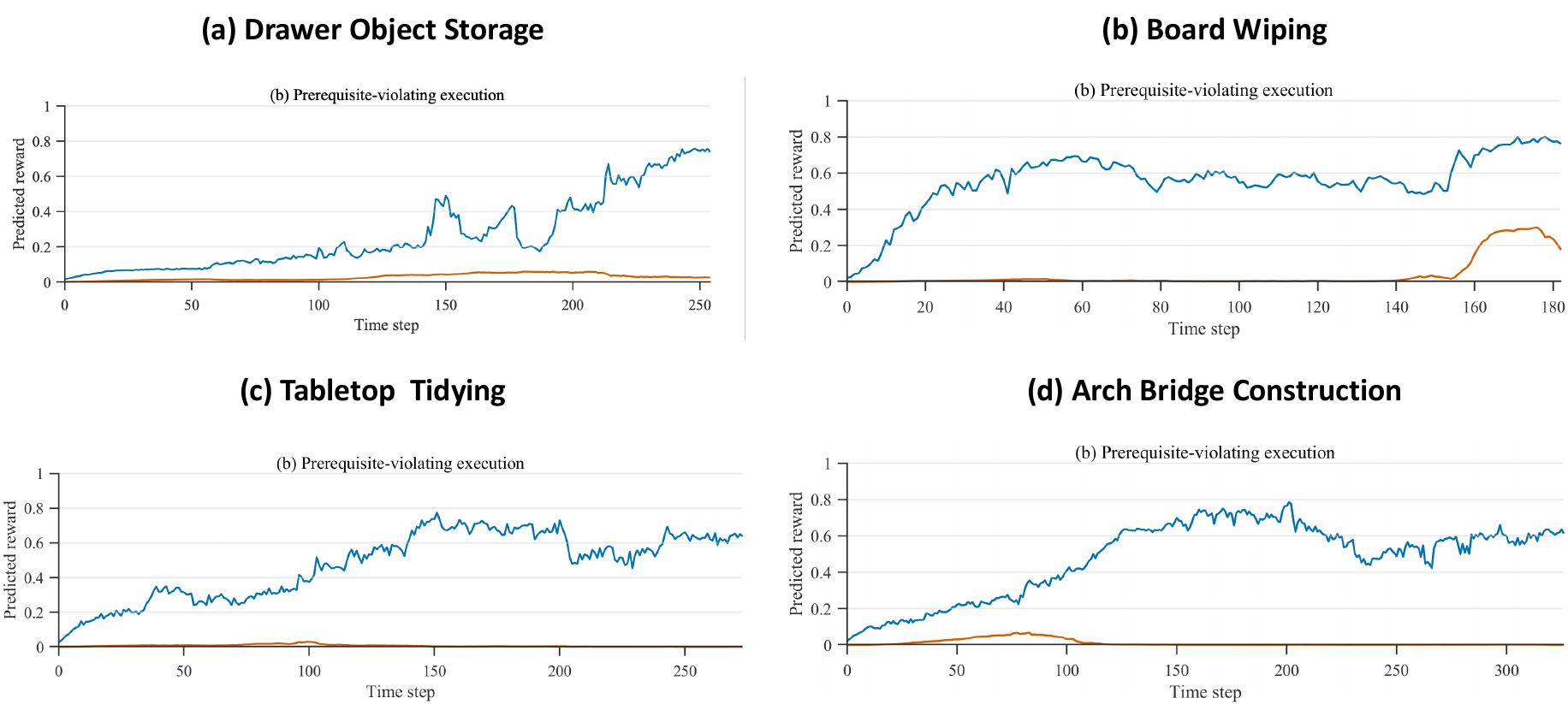}
    \caption{\textbf{Reward responses to prerequisite violations.} ReWiND (blue) and RoboLongR (orange) rewards on held-out failed rollouts containing prerequisite violations. Valid executions are shown in Fig.~\ref{fig:real_valid_reward}.}
    \label{fig:real_valid_invalid_rewards}
    \vspace{-2mm}
\end{figure}

\noindent\textbf{Component contributions.}
We remove individual components under shared demonstrations and interaction budgets to assess their contributions to downstream policy learning (Fig.~\ref{fig:ablation}). Removing stage-shuffle reduces mean success from 31.8\% to 24.3\%, a 7.5-percentage-point decrease and the largest observed drop. Removing online correction, recurrent memory, or the stage schedule yields 28.4\%, 28.9\%, and 29.8\%, respectively. These comparisons support the utility of each component within the evaluated pipeline.

\noindent\textbf{Stage scheduling.}
The no-schedule control fixes the stage input to 1 throughout PPO while retaining BC, architecture, and reward correction. The full configuration achieves a 2.0-percentage-point gain under the same interaction budget. Because initial conditioning and subsequent scheduling change, this comparison measures their combined effect and does not isolate the contribution of individual stage transitions.

\noindent\textbf{Reward refinement and stage representations.}
One, two, and three reward-training rounds yield 28.8\%, 31.8\%, and 32.0\% success, indicating that two rounds capture most of the observed improvement. Using coarse segments for stage-shuffle reduces success to 29.8\%, while using adaptive indices for both reward augmentation and policy conditioning yields 30.4\%. These results support separate stage representations: trajectory-specific partitions for reward supervision and a shared coarse vocabulary for policy conditioning.

\subsection{Real-Robot Validation}
\label{sec:real_robot}

\noindent\textbf{Platform and tasks.}
We evaluate real-world policy adaptation on a seven-degree-of-freedom Franka Panda with front and right Intel RealSense D435 cameras (Fig.~\ref{fig:real_setup}). Four tasks cover articulated-object interaction, tool use, multi-object manipulation, and assembly (Fig.~\ref{fig:real_tasks}). Drawer Object Storage requires opening the drawer, inserting the object, and closing it. Board Wiping requires retrieving a tool before cleaning. Tabletop Tidying requires stacking three cups before transferring them into a basket. Arch Bridge Construction requires positioning two supports before placing the arch.

\noindent\textbf{Training and evaluation.}
We compare RoboLongR with ReWiND, the strongest simulation baseline. Both methods use the same ten successful GELLO~\cite{wu2024gello} demonstrations per task for reward learning and BC initialization, followed by online refinement under the same interaction budget. Reward models start from their respective RLBench/Open-X checkpoints and retain their original objectives during task-specific fine-tuning. Each method is evaluated in ten trials per task with randomized placements. Success requires an object inside the fully closed drawer, a clean marked region with the tool released, a stable cup stack inside the basket, or an arch resting on upright supports, respectively. One annotator blind to the method judges complete-task success from video, and failed trials are not selectively repeated. Appendix~\ref{app:real_task} specifies workspace controls and trial records.

\begin{table}[!t]
\centering
\caption{\small \textbf{Real-robot success rates (\%).} Ten trials per task and method. Brackets give Wilson 95\% intervals.}
\label{tab:real_robot}
\small
\begin{adjustbox}{max width=\linewidth}
\begin{tabular}{lcc}
\toprule
Task & ReWiND & RoboLongR \\
\midrule
Drawer Object Storage & 20 [5.7, 51.0] & \textbf{70} [39.7, 89.2] \\
Board Wiping & 50 [23.7, 76.3] & \textbf{70} [39.7, 89.2] \\
Tabletop Tidying & 20 [5.7, 51.0] & \textbf{50} [23.7, 76.3] \\
Arch Bridge Construction & 10 [1.8, 40.4] & \textbf{40} [16.8, 68.7] \\
\midrule
Pooled (40 trials) & 25.0 [14.2, 40.2] & \textbf{57.5} [42.2, 71.5] \\
\bottomrule
\end{tabular}
\end{adjustbox}
\vspace{-2mm}
\end{table}

\noindent\textbf{Policy improvement.}
RoboLongR succeeds in 23/40 trials versus ReWiND's 10/40, giving pooled success rates of 57.5\% and 25.0\% and a 32.5-percentage-point gain (Table~\ref{tab:real_robot}). Observed success is higher on all four tasks. With ten trials per task, Wilson 95\% intervals remain wide. The pooled interval describes this fixed task mixture and does not establish per-task statistical significance.

\noindent\textbf{Reward behavior.}
We analyze separately recorded held-out rollouts excluded from reward training, BC, policy optimization, and model selection. Valid executions assess whether rewards track task progress (Fig.~\ref{fig:real_valid_reward}), while failed executions test their response to missing prerequisites (Fig.~\ref{fig:real_valid_invalid_rewards}). Both methods track valid progress, with lower raw RoboLongR scores on some tasks. Because score scales differ across independently trained models, we compare valid and failed executions within each model.

In these held-out rollouts, ReWiND assigns substantial rewards to both valid and failed executions. RoboLongR shows clearer separation, particularly for Drawer Object Storage, Tabletop Tidying, and Arch Bridge Construction. On Board Wiping, its score rises modestly near the end of the failed execution as the scene resembles completed wiping, but remains below its score for the valid execution. These examples show how prerequisite violations can leave scenes visually similar to successful outcomes.

The failed drawer rollout ends without storing the object. A premature attempt followed by successful recovery need not imply task failure. These qualitative comparisons characterize the illustrated executions and do not isolate stage-shuffle from the other components.

\section{Conclusion and Discussion}
We present \textbf{RoboLongR}, an offline-to-online framework pursuing causal consistency through history-dependent rewards for long-horizon policy adaptation. Stage-order supervision and recurrent memory encourage prerequisite-sensitive progress estimates, while stage-conditioned BC-to-RL learning and terminal-feedback correction connect reward guidance with policy experience. After pretraining on 88 RLBench tasks and Open-X data, evaluations on 18 held-out tasks and four real-robot tasks show improved complete-task success over the baselines. Reward controls probe sensitivity to tested order perturbations, and stage-shuffle contributes the largest gain in component ablations.

Limitations concern supervision and reward alignment. Synthetic shuffles can introduce boundary cues and penalize valid orders or recoveries. Whole-rollout correction can suppress valid prefixes and cause reward drift, while progress return can favor prolonged high-scoring executions over early completion. Future work targets better alignment with task success, localized correction, and broader tests of valid orders and recoveries. Controlled comparisons are needed to separate scheduling effects from initial conditioning.

\bibliographystyle{IEEEtran}
\bibliography{reference}

\onecolumn
\raggedbottom
\setlength{\intextsep}{\medskipamount}
\useRomanappendicesfalse
\appendices
\makeatletter
\let\section\@IEEEappendixsavesection
\makeatother
\section*{Appendix}

\section{Simulation Task Details}
\label{app:sim_tasks}

\noindent\begin{minipage}{\textwidth}
\subsection{Reward-Pretraining Task Inventory}
\label{app:pretrain_inventory}
Table~\ref{tab:pretrain_tasks} lists the RLBench~\cite{james2020rlbench} reward-pretraining tasks and summarizes their instructions. Dataset sizes and sampling proportions are specified in Sec.~\ref{sec:setup}.

\begin{table}[H]
\centering
\caption{\textbf{Reward-pretraining task inventory.} The 88 RLBench tasks and their instruction descriptions follow the benchmark implementations.}
\label{tab:pretrain_tasks}
{\footnotesize
\setlength{\tabcolsep}{2pt}
\renewcommand{\arraystretch}{0.92}
\begin{tabular*}{\textwidth}{@{\extracolsep{\fill}}>{\raggedright\arraybackslash}p{0.20\textwidth}>{\raggedright\arraybackslash}p{0.27\textwidth}>{\raggedright\arraybackslash}p{0.20\textwidth}>{\raggedright\arraybackslash}p{0.26\textwidth}@{}}
\toprule
\textbf{Task} & \textbf{Description} & \textbf{Task} & \textbf{Description} \\
\midrule
Basketball In Hoop & shoot the ball through the net & Play Jenga & remove the protruding block without toppling \\
Beat the Buzz & slide the ring along the pole without contact & Plug Charger In Power Supply & plug the charger into the mains \\
Change Channel & point the remote and change the channel & Pour From Cup To Cup & pour one cup's contents into the other \\
Change Clock & adjust the needles to the target time & Press Switch & flick the switch \\
Close Box & close the lid on the box & Push Button & push down the specified button \\
Close Door & grip the handle and pull the door shut & Push Buttons & push the buttons in the specified order \\
Close Drawer & slide the specified drawer shut & Put Bottle In Fridge & open the fridge and put the bottle in \\
Close Fridge & shut the fridge door & Put Groceries In Cupboard & put the specified item in the cupboard \\
Close Grill & lower the grill cover by its handle & Put Knife In Knife Block & slide the knife into its slot in the block \\
Close Jar & screw the lid onto the specified jar & Put Knife On Chopping Board & move the knife from block to chopping board \\
Close Laptop Lid & shut the laptop lid & Put Money In Safe & put the money on the specified safe shelf \\
Close Microwave & close the microwave door & Put Plate In Colored Dish Rack & leave the plate between the specified spokes \\
Weighing Scales & put the specified pepper on the scales & Put Rubbish In Bin & drop the rubbish into the bin \\
Get Ice From Fridge & fill the cup with ice from the fridge & Put Toilet Roll On Stand & slide the toilet roll onto its stand \\
Hang Frame On Hanger & hang the picture frame on the hanger & Put Tray In Oven & open the oven and slide the tray in \\
Hit Ball With Queue & pot the ball into the goal with the cue & Put Umbrella In Umbrella Stand & place the umbrella into its stand \\
Hockey & push the ball into the goal with the stick & Reach And Drag & drag the cube onto the target with the stick \\
Insert Onto Square Peg & slide the ring onto the specified spoke & Reach Target & touch the specified sphere with the gripper \\
Insert Usb In Computer & slide the USB stick into the slot & Remove Cups & take one cup off the holder onto the table \\
Lamp Off & press the switch to turn the lamp off & Scoop With Spatula & scoop the cube up with the spatula and lift \\
Lamp On & press the switch to turn the lamp on & Screw Nail & screw in the nail with the screwdriver \\
Lift Numbered Block & lift the block with the target number & Slide Block To Target & push the block onto the target \\
Light Bulb In & screw the bulb from its holder into the lamp & Slide Cabinet Open & slide the specified cabinet half open \\
Light Bulb Out & unscrew the bulb and leave it in the stand & Stack Wine & slide the bottle onto the wine rack \\
Meat Off Grill & take the specified meat off the grill & Straighten Rope & pull each rope end until it is straight \\
Meat On Grill & put the specified meat on the grill & Sweep To Dustpan & brush the dirt into the dustpan \\
Move Hanger & move the hanger to the other rack & Take Cup Out From Cabinet & open the cabinet and retrieve the cup \\
Open Box & open the box lid & Take Frame Off Hanger & slide the photo off the hanger onto the table \\
Open Door & grip the handle and push the door open & Take Item Out Of Drawer & open the drawer and take the cube out \\
Open Drawer & pull the specified drawer open & Take Lid Off Saucepan & lift the lid off the saucepan by its handle \\
Open Fridge & slide the fridge door open & Take Money Out Safe & take the money off the shelf onto the table \\
Open Grill & raise the grill cover by its handle & Take Off Weighing Scales & remove the pepper from the scales \\
Open Jar & unscrew the lid off the specified jar & Take Plate Off Colored Dish Rack & remove the plate from the specified spokes \\
Open Microwave & pull the microwave door open & Take Shoes Out Of Box & open the box and put the shoes on the table \\
Open Oven & pull the oven door open by its handle & Take Toilet Roll Off Stand & slide the roll off its stand onto the table \\
Open Washing Machine & pull the washing machine door open & Take Tray Out Of Oven & open the oven and take the tray out \\
Open Window & unlock the handle and push the window open & Take Umbrella Out Of Umbrella Stand & lift the umbrella out of its stand \\
Open Wine Bottle & unscrew and remove the bottle cap & Take Usb Out Of Computer & remove the USB stick from its port \\
Phone On Base & put the handset back on its base & Toilet Seat Down & close the toilet seat by its lid \\
Pick And Lift & lift the specified block to the target & Toilet Seat Up & lift the toilet seat upright \\
Pick And Lift Small & lift the specified small object to the target & Turn Oven On & rotate the leftmost knob to turn the oven on \\
Pick Up Cup & grasp and lift the specified cup & Turn Tap & grasp and turn the specified tap handle \\
Place Hanger On Rack & place the clothes hanger on the rack & Tv On & point the remote and turn the television on \\
Place Shape In Shape Sorter & put the shape into its matching slot & Unplug Charger & pull the charger out of its socket \\
\bottomrule
\end{tabular*}
}

\end{table}
\end{minipage}
\par\medskip

\noindent\begin{minipage}{\textwidth}
\subsection{Evaluation-Task Prerequisites}
\label{app:evaluation_tasks}
Table~\ref{tab:lh_tasks} lists prerequisite relations and valid alternative orders for each evaluation task. Groups are defined before policy training using RLBench success conditions and physical task structure, independently of method performance.

\begin{table}[H]
\centering
\caption{\textbf{Evaluation-task prerequisites.} Relations and valid alternative orders for the 18 evaluation tasks. $\prec$ denotes precedence.}
\label{tab:lh_tasks}
{\scriptsize
\setlength{\tabcolsep}{2pt}
\renewcommand{\arraystretch}{0.96}
\begin{tabular*}{\textwidth}{@{\extracolsep{\fill}}>{\raggedright\arraybackslash}p{0.20\textwidth}>{\raggedright\arraybackslash}p{0.45\textwidth}>{\raggedright\arraybackslash}p{0.27\textwidth}@{}}
\toprule
\textbf{Task} & \textbf{Prerequisite structure} & \textbf{Valid alternative orders} \\
\midrule
\multicolumn{3}{l}{\textcolor{StrongGroup}{\textit{Strong dependency}}} \\
Put Item In Drawer & open drawer $\prec$ place item $\prec$ close drawer & none across the three subtasks \\
Empty Dishwasher & open door $\prec$ pull out rack $\prec$ remove plate & none \\
Slide Cabinet Open \& Place Cups & open cabinet $\prec$ place cups & cup order, when several cups are used \\
Water Plants & grasp can $\prec$ approach plant $\prec$ pour & none \\
Wipe Desk & grasp sponge $\prec$ contact desk $\prec$ remove dirt & order among independent dirt patches \\
Block Pyramid & bottom layer $\prec$ middle layer $\prec$ top layer & placements within one layer \\
Stack Blocks & lower block $\prec$ block supported by it & orders preserving all support relations \\
Stack Cups & support cup $\prec$ cup placed above it & orders preserving all support relations \\
Stack Chairs & support chair $\prec$ chair placed above it & orders preserving all support relations \\
\midrule
\multicolumn{3}{l}{\textcolor{WeakGroup}{\textit{Weak dependency}}} \\
Place Cups & independent placement for each cup & any cup order \\
Put All Groceries In Cupboard & independent placement for each item & any item order \\
Setup Checkers & independent placement for each checker & any checker order \\
Setup Chess & independent placement for each piece & any piece order \\
Set The Table & independent placement for each tableware item & any tableware order \\
Put Books On Bookshelf & independent placement for each book & any book order \\
Put Shoes In Box & open lid $\prec$ place either shoe & either shoe order \\
Solve Puzzle & independent insertion for each puzzle piece & any piece order \\
Empty Container & independent transfer for each object & any object order \\
\bottomrule
\end{tabular*}
}

\end{table}
\end{minipage}
\par\medskip

\begin{multicols}{2}
\small
\raggedcolumns
\section{Additional Implementation Details}
\label{app:impl}

\subsection{Reward Inputs and Recurrent State}
\label{app:reward_inputs}
The reward model in Sec.~\ref{sec:reward_learning} encodes each camera view independently and mean-pools the view features before projection. Images are resized and center-cropped to $224\times224$. The single-layer recurrent module starts each episode from a zero state. Semantic subtask labels support augmentation construction and policy conditioning, but are not inputs to the reward model.

\subsection{Baseline Objectives}
\label{app:baselines}
\noindent\textbf{ReWiND.} We use the released architecture and default hyperparameters~\cite{zhang2025rewind} within the shared policy-learning protocol of Sec.~\ref{sec:setup}.

\noindent\textbf{Progress+Mono.} A stage-classification head $q_\psi(z_t\mid o_{1:t},\ell)$ supplements progress regression. For a trajectory of unpadded length $T$, the objective is
\begin{equation}
\label{eq:pm_objective}
\begin{aligned}
\mathcal{L}_{\mathrm{P+M}}
={}&\mathcal{L}_{\mathrm{pos}}
+\lambda_{\mathrm{stage}}\operatorname{CE}(q_\psi,z)\\
&+\frac{\lambda_{\mathrm{mono}}}{T-1}
\sum_{t=2}^{T}\max(0,\hat z_{t-1}-\hat z_t),
\end{aligned}
\end{equation}
where $\hat z_t=\sum_{k=1}^{K_c}kq_\psi(k\mid o_{1:t},\ell)$ is the expected coarse-stage index and $\lambda_{\mathrm{stage}}=\lambda_{\mathrm{mono}}=1$. With $\mathcal M$ denoting valid, unpadded steps, the two supervised terms are
\begin{equation}
\label{eq:pm_progress}
\mathcal{L}_{\mathrm{pos}}
=\frac{1}{|\mathcal M|}\sum_{t\in\mathcal M}
\left(R_\phi(o_{1:t},\ell)-y_t\right)^2,
\end{equation}
\begin{equation}
\label{eq:pm_stage}
\operatorname{CE}(q_\psi,z)
=-\frac{1}{|\mathcal M|}\sum_{t\in\mathcal M}
\log q_\psi(z_t\mid o_{1:t},\ell).
\end{equation}
Here $z_t$ denotes the coarse policy-stage label. The monotonicity term penalizes decreases in the expected stage index between consecutive valid steps. Positive regression uses no shuffled negatives.

\subsection{Randomization and Hyperparameter Selection}
\label{app:randomization}
Each seed controls network initialization, demonstration sampling, simulator randomization, and PPO sampling. Hyperparameters are selected on the reward-pretraining validation split and fixed without access to evaluation-task results. All methods use PyTorch and the same RLBench software environment.

\subsection{Real-Robot Controls and Trial Records}
\label{app:real_task}
The Panda uses a parallel-jaw gripper. Camera inputs use only the RGB streams from the two RealSense cameras. The workspace, camera poses, objects, controller, and initial-pose sampling regions are fixed across methods. Initial object poses are sampled uniformly from predefined task-specific workspace regions.

Controller resets, safety stops, human interventions, and hardware failures are recorded separately. Trial outcomes follow the complete-task success criteria in Sec.~\ref{sec:real_robot}. Representative execution failures include missed grasps, premature releases, collisions, and inaccurate placements.

\section{Additional Reward Analysis}
\label{app:more_analysis}

\subsection{Adaptive Partition Statistics}
\label{app:adaptive_partition}
Table~\ref{tab:fallback} summarizes the frequency of detected curvature-based boundaries and the mean number of adaptive segments across task categories. Boundary detection is most frequent in opening and closing tasks and least frequent in composition tasks. Segment counts vary by trajectory and can exceed the uniform fallback count in Sec.~\ref{sec:reward_learning}.

\begin{table}[H]
\centering
\caption{\textbf{Adaptive partition statistics.} Frequency of detected curvature-based boundaries and mean adaptive segment counts.}
\label{tab:fallback}
\small
\begin{adjustbox}{max width=\linewidth}
\begin{tabular}{lcc}
\toprule
\textbf{Category} & \textbf{Curvature-based (\%)} & \textbf{Mean adaptive stages} \\
\midrule
Pick-and-place & 78 & 4.6 \\
Opening and closing & 91 & 3.8 \\
Insertion and alignment & 84 & 4.1 \\
Composition & 52 & 6.2 \\
\bottomrule
\end{tabular}
\end{adjustbox}

\end{table}

\subsection{Evaluation Swap Distances}
\label{app:permutation_distance}
Evaluation swaps exchange intact segments at distances one (adjacent), two (one intermediate segment), and at least three (wider separations). Adjacent swaps can overlap the training corruption family.

Aggregate diagnostics follow Sec.~\ref{sec:q2} and Table~\ref{tab:causal_sensitivity}. Distance can affect both execution-history mismatch and boundary discontinuities, so it does not isolate sensitivity to either factor.
\end{multicols}

\end{document}